\pdfoutput=1

\documentclass[11pt]{article}

\usepackage{emnlp2021}

\usepackage{times}
\usepackage{latexsym}
\usepackage{amsmath, amssymb}
\usepackage[pdftex]{graphicx}
\usepackage{enumitem}
\usepackage{tabularx}
\usepackage{bbding}
\usepackage{multirow}
\usepackage{listings}
\usepackage{color}
\usepackage{tablefootnote}

\usepackage[T1]{fontenc}

\usepackage[utf8]{inputenc}

\usepackage{microtype}

\def\code#1{\texttt{#1}}

\title{Code Generation for Unknown Libraries via Reading API Documentations}

\author{Koki Washio \\
  The University of Tokyo \\
  \texttt{kwashio@is.s.u-tokyo.ac.jp} \\\And
  Yusuke Miyao \\
  The University of Tokyo \\
  \texttt{yusuke@is.s.u-tokyo.ac.jp} \\}

\begin{document}
\maketitle
\begin{abstract}
Open-domain code generation is a challenging problem because the set of functions and classes that we use are frequently changed and extended in programming communities.
We consider the challenge of code generation for unknown libraries without additional training.
In this paper, we explore a framework of code generation that can refer to relevant API documentations like human programmers to handle unknown libraries.
As a first step of this direction, we implement a model that can extract relevant code signatures from API documentations based on a natural language intent and copy primitives from the extracted signatures.
Moreover, to evaluate code generation for unknown libraries and our framework, we extend an existing dataset of open-domain code generation and resplit it so that the evaluation data consist of only examples using the libraries that do not appear in the training data. 
Experiments on our new split show that baseline encoder-decoder models cannot generate code using primitives of unknown libraries as expected.
In contrast, our model outperforms the baseline on the new split and can properly generate unknown primitives when extracted code signatures are noiseless.
\end{abstract}

\section{Introduction} \label{sec: Introduction}

Semantic parsing, the task of mapping natural language into a formal representation that is executable for machines, is important for achieving seamless interaction between humans and machines.
Formal representation in semantic parsing includes domain-specific languages for specific applications \citep{zelle-and-mooney1996, berant-etal-2013-semantic, quirk-etal-2015-language} and knowledge base queries such as SQL \citep{atis-pepar}.
In addition to these, there has also been a lot of studies about code generation, the task of mapping natural language intents into general-purpose programming languages such as Python.
Recent works in semantic parsing and code generation have developed supervised neural encoder-decoder models to achieve good performance \citep{dong-lapata-2016-language, liang-etal-2017-neural, ling-etal-2016-latent, rabinovich-etal-2017-abstract, yin-neubig-2017-syntactic, yin-etal-2018-structvae, dong-lapata-2018-coarse, iyer-etal-2018-mapping, yin-neubig-2019-reranking}.

However, open-domain code generation with ordinary supervised models has difficulties.
Although supervised neural models can achieve good performance when training data cover all primitives, which are symbols representing functions, classes, etc., they cannot handle the unknown primitives that do not appear in training data \cite{herzig-berant-2018-decoupling}.
In code generation, the number of primitives of general-purpose programming languages is huge and it is frequently extended as libraries by daily programming.
Thus, a model needs to handle functions and classes that do not appear in the training and pretraining data.

\begin{figure}
    \centering
    \includegraphics[width=7.0cm]{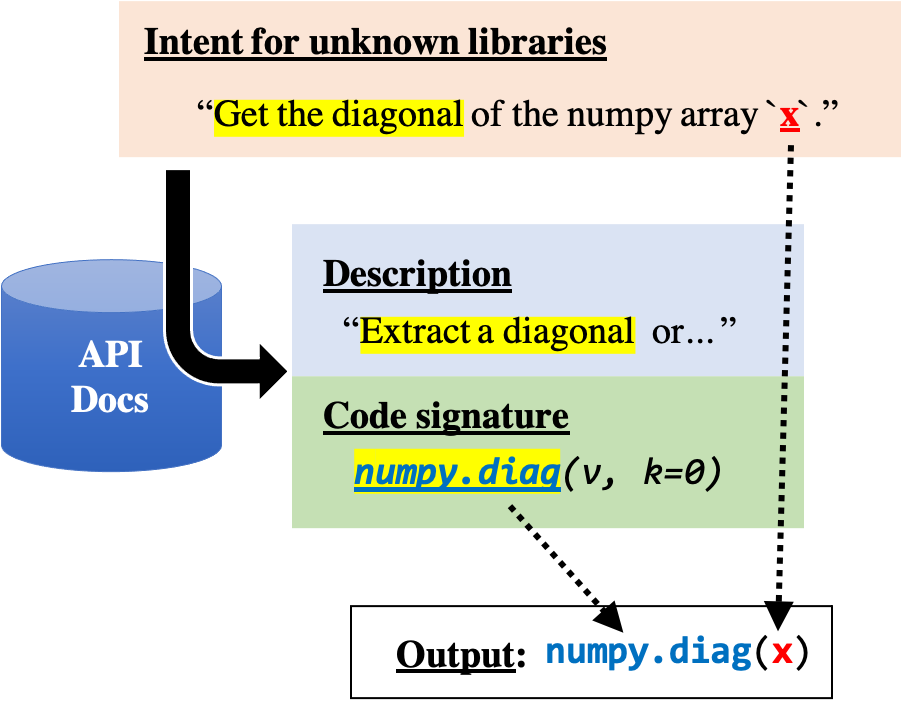}
    \caption{Our framework for code generation via exploiting API documentations.}
    \label{fig:framework}
\end{figure}

In this paper, we explore a framework of code generation that can adapt to unknown libraries by exploiting API documentations.
Figure \ref{fig:framework} shows the overview of our framework in Python code generation.
Supposing that Numpy used in the figure is an unknown library, the ordinary supervised framework cannot generate the output snippet because the primitive \code{diag} is not contained in the output vocabulary.
In contrast, human programmers can search in the API documentation based on their intent and write a code by referring to the code signatures even if they have never used those functionalities before.
In other words, they can do zero-shot adaptation to unknown formal representations through document understanding.
Utilizing API documentations can be also regarded as incorporating natural language understanding, such as question answering and reading comprehension \citep{chen-etal-2017-reading}, into code generation to achieve the zero-shot adaptation.

From a practical point of view, if a model can handle unknown libraries by reading documents like human programmers, we do not have to collect additional training examples for new libraries or re-train our model.
All we have to do for new libraries is to add the corresponding API documentations to document sets for the model.
Moreover, if we know in advance which libraries we will not use, we can adjust the output space just by removing them from the document set.

As a first step to realizing this framework in Figure \ref{fig:framework}, we implement an encoder-decoder model that can copy primitives from code signatures in API documentations.
Given an intent, our model first extracts relevant pairs of code signatures and natural language descriptions from the document set, like open-domain question answering.
Then, the model encodes the pairs into their hidden representations with the attention mechanism between the signatures and descriptions.
During decoding, our model copies the primitives from the extracted signatures based on the signature representations by the copy mechanism \citep{gu-etal-2016-incorporating, merity2016pointer, see-etal-2017-get}.

To evaluate code generation for adapting to unknown libraries by reading API documentations, we extend and re-split the CoNaLa dataset \citep{yin-etal-2018-learning}, which is a parallel corpus of natural language intents and Python snippets.
For each example, we enumerate the primitives that appear in the Python snippet and annotate the corresponding code signatures and natural language descriptions extracted from the API documentations.
Then, we re-split the datasets to evaluate zero-shot adaptation for unknown libraries.
In our new data split, the training set includes examples that only use built-in functions and built-in libraries, while the development set and test set consist of those that use third-party libraries. This allows us to evaluate code generation for unknown libraries.

Our experiments show that the performance of models on our new split was much lower than on the random split and the ordinary supervised baseline could generate no unknown primitives as we expected.
In contrast, our model that outperformed the baseline could copy unknown primitives from API documentations.

The main contributions of our paper as follows:
(1) we construct a dataset to evaluate code generation for unknown libraries by extending an existing dataset;
(2) we propose a framework of code generation with reading API documents to adapt to unknown formal representations and libraries without additional training.
We will make the code and data public for future research.

\section{Code Generation Problem} \label{sec: Code Generation Problem}
Code generation is the task of generating a code snippet $\bar{c} = (c_1, \ldots, c_{|\bar{c}|})$ given a natural language intent $\bar{x} = (x_1, \dots, x_{|\bar{x}|})$.
In this task, we estimate $p(\bar{c}|\bar{x})$, the conditional probability of a snippet $\bar{c}$ given an intent $\bar{x}$ as follows:

\begin{equation}
    p(\bar{c}|\bar{x}) = \prod_{t=1}^{|\bar{c}|} p(c_t | \bar{c}_{<t}, \bar{x})
\end{equation}
where $\bar{c}_{<t} = (c_1, \ldots, c_{t-1})$.
To model this probability distribution, previous works used supervised models based on training data that consist of pairs of intents $\bar{x}$ and snippets $\bar{c}$ \citep{ling-etal-2016-latent, rabinovich-etal-2017-abstract, yin-neubig-2017-syntactic, dong-lapata-2018-coarse, iyer-etal-2018-mapping}.
This ordinary supervised framework achieves good results when the distributions of training and test data are equal.

Models in this framework, however, cannot generate code snippets that include primitives that do not appear in the training data.
The set of primitives such as functions and classes that we want to use are frequently changed and extended by developing new libraries in programming communities.
In this framework, we have to recollect training data and retrain our model to handle such daily changes and extensions.
To make matters worse, the collection of training data for code generation is very costly.


\section{Model} \label{sec: Model}
We explore the framework of code generation that can adapt to unknown libraries by understanding API documentations.
An API documentation is a human-readable resource that describes the usages and purposes of functions and classes.
It mainly consists of \textit{code signatures} and \textit{natural language descriptions}\footnote{There is a kind of API documentations that is more structured or contains additional information, such as usage examples. We leave using such additional information for future work.}.
Code signatures provide a prototype of the usages, and their natural language descriptions explain the purposes and functionalities.
Based on this information, programmers can correctly use unknown primitives.

Our key insight is as follows: 
when generating a snippet given an intent, a model can adapt to unknown libraries like human programmers if it can refer to the appropriate components in the code signatures in the relevant API documentations.
To realize this framework, we propose a model that can extract relevant code signatures from documentations given an intent and copy symbols from the extracted signatures.
Our model consists of the three modules:
(1) \textbf{API retriever} extracts relevant pairs of code signatures and descriptions from a document set by inferring relevancy of a given intent and descriptions like open-domain question answering;
(2) \textbf{API reader} encodes each extracted signature with aligning each component to the corresponding description by attention mechanism;
(3) \textbf{Code generator} is an encoder-decoder model with the copy mechanism that can copy symbols from encoded signatures passed from the API reader.
The following sections describe each module.

\subsection{API Retriever} \label{subsec: API Retriever}
This module takes an intent $\bar{x}$ as input and calculates relevant scores for each pair $a = (\bar{s}, \bar{d})$ of a code signature $\bar{s} = (s_1, \ldots, s_{|\bar{s}|})$ and description $\bar{d} = (d_1, \ldots, d_{|\bar{d}|})$ in a document set $D = \{a_1, \ldots, a_{|D|}\}$.
We formulate the scoring function $f_r$ as follows:
\begin{equation} \label{eq: vectorizer}
    f_r(\bar{x}, a) = \frac{\mathbf{v}_{\bar{x}} \cdot \mathbf{v}_{\bar{d}}}{|\mathbf{v}_{\bar{x}}| |\mathbf{v}_{\bar{d}}|}
\end{equation}
where $\cdot$ is the dot product; $\mathbf{v}_{\bar{x}}$ and $\textbf{v}_{\bar{d}}$ are vectroized intent $\bar{x}$ and description $\bar{d}$ of $a$, respectively.
This function calculates the cosine similarity of the vectorized intent and description.
Since this relevance scoring is similar to the scoring between queries and documents in information retrieval and question answering, the techniques in those fields such as the tf-idf method can be exploited.
According to the assigned scores, top $k$ signature-description pairs are extracted as $D^\prime = \{a^\prime_1, \ldots, a^\prime_k \}$ and passed to the following API reader.

\subsection{API Reader} \label{subsec: API Reader}
The API reader encodes each extracted code signature $\bar{s} = (s_1, \ldots, s_{|\bar{s}|})$ in $D^\prime$ as aligned hidden representations of the corresponding natural language description $\bar{d} = (d_1, \ldots, d_{|\bar{d}|})$.
To handle out-of-vocabulary (OOV) symbols in code signatures, we replace the tokens outside the parentheses with a special token \texttt{FUNC}.
By aligning the hidden representations of the description to each element of the signature using the attention, the following code generator can recognize what each symbol in the signature represents when deciding whether to copy $s_i$ or not.

This module has two neural sequence encoders for $\bar{s}$ and $\bar{d}$, respectively.

\small
\begin{eqnarray}
    (\mathbf{s}^{\prime}_1, \ldots, \mathbf{s}^{\prime}_{|\bar{s}|}) &=& \mathrm{Encoder_{SG}} (\mathbf{s}_1, \ldots, \mathbf{s}_{|\bar{s}|}) \\
    \label{eq: encoder_nl}
    (\mathbf{d}^{\prime}_1, \ldots, \mathbf{d}^{\prime}_{|\bar{d}|}) &=& \mathrm{Encoder_{NL}} (\mathbf{d}_1, \ldots, \mathbf{d}_{|\bar{d}|})
\end{eqnarray}
\normalsize
where $\mathrm{Encoder_{SG}}$ and $\mathrm{Encoder_{NL}}$ are neural sequence encoders, such as the long short-term memory (LSTM);
$\mathbf{s}_i$ and $\mathbf{d}_j$ are the embeddings of each token in the signature and description;
$\mathbf{s}^{\prime}_i$ and $\mathbf{d}^{\prime}_j$ are the hidden representations from the encoders corresponding to $s_i$ and $d_j$, respectively.

Then, this module calculates attention scores to align each component of the signature with the description.
The attention score $\alpha_{ij}$ between $s_i$ and $d_j$ is calculated as follows:
\begin{equation}
    \alpha_{ij} = \mathrm{Att}_{doc} (\mathbf{s}^{\prime}_i, \mathbf{d}^{\prime}_j)
\end{equation}
where $\mathrm{Att}_{*}$ is a trainable scalar function, such as the bilinear attention.
Based on $\alpha_{ij}$, we assign the hidden representation of the description to $s_i$.
\begin{equation}
    \label{eq: singature representations}
    \mathbf{h}^{s}_{i} = \sum_{j = 1}^{|\bar{d}|} \frac{e^{\alpha_{ij}}}{\sum_{k=1}^{|\bar{d}|} e^{\alpha_{ik}}} \mathbf{d^\prime}_j
\end{equation}
where $\mathbf{h}^{s}_{i}$ is a representation of the natural language description corresponding to $s_i$.
During decoding, to consider this representation by the copy mechanism allows the model to refer to the relevant API documentations.

\subsection{Code Generator} \label{subsec: Code Generator}
The code generator is an encoder-decoder model with the copy mechanism that generates the snippet $\bar{c}$ given the natural language intent $\bar{x}$ and the encoded signatures passed from the API reader.
In this paper, we extend the CopyNet \citep{gu-etal-2016-incorporating} so that the model can refer to the code signatures extracted from the document set.

\subsubsection*{Encoding and Decoding}
This module uses the $\mathrm{Encoder_{NL}}$ in Equation (\ref{eq: encoder_nl}) to encode the natural language intent $\bar{x}$ as follows:
\begin{equation}
    \label{eq: intent representation}
    (\mathbf{h}^x_1, \ldots, \mathbf{h}^x_{|\bar{x}|}) = \mathrm{Encoder_{NL}} (\mathbf{x}_1, \ldots, \mathbf{x}_{|\bar{x}|})
\end{equation}
where $\mathbf{h}^x_i$ is a hidden representation of the intent corresponding to $x_i$ and $\mathbf{x}_i$ is the embedding of $x_i$.
These representations are used in calculating the decoder hidden states and the copying scores.
The last hidden representation $\mathbf{h}^x_{|\bar{x}|}$ is used also as the initial state of the decoder.

In the decoding process, this module generates the snippet symbol $c_t$ based on the normalized generation score $q_g$ and copying score $q_c$ as follows:

\small
\begin{eqnarray}
    p(c_t | \mathbf{z}_{t}, \bar{c}_{<t}, \bar{x}, D^{\prime}) &=& 
    q_g(c_t | \mathbf{z}_{t}, \bar{c}_{<t}, \bar{x}, D^{\prime}) \\
    &+& q_c(c_t | \mathbf{z}_{t}, \bar{c}_{<t}, \bar{x}, D^{\prime})
\end{eqnarray}
\normalsize
where $\mathbf{z}_t$ is the hidden state of the decoder at step $t$.
Here, we assume a vocabulary for output snippets $V = \{v_1, \ldots, v_{|V|}\}$ and a special symbol $\mathrm{UNK}$ for OOV elements.
We define $M$ as the set of all the unique symbols in the intent $\bar{x}$ and the extracted signatures $\{ \bar{s} | (\bar{s}, \bar{d}) \in D^\prime \}$.
In addition, we define $H = \{ \mathbf{h}^x_i \} \cup \bigcup \{ \mathbf{h}^s_j | (\bar{s}, \bar{d}) \in D^\prime \}$ as the set of all hidden representations of the intent and the extracted signatures from Equation (\ref{eq: singature representations}, \ref{eq: intent representation}), and $H_m \subseteq H$ as the set of the hidden representation corresponding to the symbol $m \in M$.
The $q_g$ and $q_c$ are defined as follows:

\begin{equation}
    q_g(c_t | \cdot) = 
    \begin{cases}
        \displaystyle
        \frac{1}{Z} e^{\psi_g(c_t)}, & c_t \in V \\
        0, & c_t \in M \cap \overline{V} \\
        \displaystyle
        \frac{1}{Z} e^{\psi_g(\mathrm{UNK})}, & c_t \notin M \cup V
    \end{cases}
\end{equation}
\begin{equation}
    q_c(c_t | \cdot) = 
    \begin{cases}
        \displaystyle
        \frac{1}{Z} e^{\psi_c (c_t)}, & c_t \in M \\
        0, & \mathrm{otherwise}
    \end{cases}
\end{equation}
where $\psi_g$ and $\psi_c$ are score functions for the generation and copying, respectively; $Z$ is the normalization term defined as $Z = \sum_{v \in V \cup \{\mathrm{UNK}\}} e^{\psi_g(v)} + \sum_{m \in M} e^{\psi_c(m)}$.

The scoring functions $\psi_g$ is defined as follows:
\begin{eqnarray}
    \psi_g(c_t = v_i) = (\mathbf{W}_g \mathbf{z}_t + \mathbf{b}_z)_i
\end{eqnarray}
where $v_i$ is the $i$-th item of the list of the elements from $V \cup {UNK}$; $\mathbf{W}_{*}$ and $\mathbf{b}_{*}$ are weight matrices and bias vectors, respectively.
This is the ordinary scoring function in the generic RNN decoder model.
On the other hand, $\psi_c$ is defined as follows:

\begin{eqnarray}
    \psi_c(c_t = m_j) = \sum_{\mathbf{h} \in H_{m_j}} f_c (\mathbf{h}, \mathbf{z}_t) \\
    f_c (\mathbf{h}, \mathbf{z}_t) = \mathrm{tanh}(\mathbf{W}_c \mathbf{h} + \mathbf{b}_c)^\top \mathbf{z}_t
\end{eqnarray}
where $\mathrm{tanh}$ is the hyperbolic tangent activation function.
The function $\psi_c$ calculates scores for copying the symbol $m_j \in M$ based on the hidden state of the decoder and the representations of the encoded intents or signatures.
This score function is extended from the original one of \citet{gu-etal-2016-incorporating} that only considers the input sequence such that it can also consider the extracted code signatures.

\subsubsection*{Decoder State Update}
The code generator updates its decoder state $\mathbf{z}_t$ based on the previous state $\mathbf{z}_{t-1}$, the predicted symbol $c_{t-1}$, and the attention to $H$.
Our model uses the attentive read and selective read following the original CopyNet as follows:

\begin{eqnarray}
    \mathbf{z}_t = f_{\mathrm{RNN}} (\mathbf{z}_{t-1}. \mathbf{i}_t) \\ 
    \mathbf{i}_t = \mathbf{W}_z [
        \mathbf{c}_{t-1} \oplus \mathbf{l}_t \oplus \mathbf{u}_t
    ] + \mathbf{b}_z 
\end{eqnarray}
where $f_{\mathrm{RNN}}$ computes the hidden state given the previous hidden state $\mathbf{z}_{t-1}$ and the current input representation $\mathbf{i}_t$ along with the decoder RNN architecture such as LSTM; 
$\oplus$ is the vector concatenation;
$\mathbf{c}_{t-1}$ is the embedding of $c_{t-1}$.
The vector $\mathbf{l}_t$ and $\mathbf{u}_t$ are the context vectors from the attentive read and selective read on $H$ calculated as follows:

\begin{eqnarray}
    \mathbf{l}_t = \sum_{\mathbf{h}_i \in H} \frac{e^{\beta_i}}{\sum_k e^{\beta_k}}\mathbf{h}_i  \\
    \mathbf{u}_t = \sum_{\mathbf{h}_j \in H} \frac{e^{\gamma_j}}{\sum_k e^{\gamma_k}} \mathbf{h}_j 
\end{eqnarray}
where $\beta_i$ and $\gamma_j$ are the attention scores for the attentive read and selective read defined as follows:
\begin{align}
    \beta_i = \mathrm{Att_{dec}} (\mathbf{h}_i, \mathbf{z}_t) \\
    \gamma_j = \begin{cases}
        \displaystyle
        \frac{1}{K} f_c(\mathbf{h}_j, \mathbf{z}_{t-1}),  & \mathbf{h}_j \in H_{c_{t-1}} \\
        0, & \mathrm{otherwise}
    \end{cases}
\end{align}
where $K$ is the normalization term defined as $\sum_{\mathbf{h}_k \in H_{c_{t-1}}} f_c (\mathbf{h}_k, \mathbf{z}_{t-1})$;
$\beta_i$ for the attentive read is the oridinary attention score to the set of the representation $H$;
$\gamma_j$ is the attention score only to $H_{c_{t-1}}$, the representations that is used for the previous copying.

\section{Library-Split Dataset} \label{sec: Libary-Split Dataset}
Existing datasets for code generation \citep{django, ling-etal-2016-latent, yin-etal-2018-learning, iyer-etal-2018-mapping, agashe-etal-2019-juice} do not assume the setting, where a model needs to generate snippets using unknown libraries.
To evaluate that setting and our framework in Section \ref{sec: Model}, we extend and resplit the CoNaLa dataset \citep{yin-etal-2018-learning}, which is a parallel corpus of natural language intents and Python snippets.

To annotate correct code signatures and descriptions for each example in the CoNaLa, we hired as annotators three students at the computer science department in a university who are familiar with Python.
First, they enumerated primitives in each snippet.
Then, they searched in API documentations the signature-description pairs that correspond to the enumerated primitives, and extracted those pairs as much as possible.
For descriptions, the annotators extract the first paragraphs.
After this process, one of the authors double-checked the results.
In this process, we exclude examples that are not Python snippets.
For the details of annotation decisions, please refer to Appendix A.

Based on these annotation results and intents, we split the dataset so that the training set includes examples whose snippets only use built-in functions and built-in libraries of Python, while the development and test set contains those that refer to third-party libraries.
By using this library-split dataset and the signature-description annotations, we can evaluate code generation for unknown libraries and our framework.
Table \ref{tab: dataset statistics} shows the statistics of the random split and library split.\footnote{For the random split, we preserve the original test set and split the training data into the train set and development set.}
Note that OOV primitives on the table are the symbols in the tokenized snippets that are contained neither in $V$ constructed from the training set nor in the corresponding intents.
These primitives cannot be generated by the ordinary supervised encoder-decoder models.
Our library split contains about six times more examples that have OOV primitives in the development set than the random split.

\begin{table}[]
\centering
\begin{tabular}{l|cc}
\hline
                & \multicolumn{1}{l}{Random} & \multicolumn{1}{l}{Library} \\ \hline
\# Train        & 2176                       & 2146                              \\
\# Dev          & 200                        & 200                               \\
\# Test         & 499                        & 529                               \\ \hline
\# OOV primitives  & 45                         & 375                               \\
\# OOV examples & 28                         & 166                               \\
\% OOV examples & 14                         & 83                                \\ \hline
\end{tabular}
\caption{\label{tab: dataset statistics}
The dataset statistics of the random split and our library split. The table shows the number of examples in the train/development/test set and the number of OOV primitives and examples that contain OOV primitives in the development set.
}
\end{table}

\section{Expriments} \label{sec: Experiments}

Using the dataset in Section \ref{sec: Libary-Split Dataset}, We conduct experiments to investigate the performance degradation on the library-split dataset and to evaluate our framework that the model can refer to API documentations.
Following the previous works using the CoNaLa dataset \citep{yin-neubig-2019-reranking, xu-etal-2020-incorporating}, we use the BLEU metric\footnote{We used the implementation from the official baseline code in \url{https://github.com/conala-corpus/conala-baseline/}} to measure performance.

\subsection{Document Set and Retriever Selection}

As a document set $D$, we use not only the annotated signature-description pairs in the dataset but also those extracted from the Python API documentations\footnote{We used the preprocessed Python documentations from \cite{xu-etal-2020-incorporating}. This contains variations of the same primitives. We exclude these variations by preserving only the longest one.} to approximate a real-world situation, resulting in $|D|=8189$.
This document set does not exactly represent the actual situation because it does not cover all code signatures of each library used in the data.
Using this document set corresponds to the situation where a user knows the candidate signatures to use.

We select a model of the API retriever to vectorize intents and descriptions in Equation (\ref{eq: vectorizer}) from the following three methods:
\begin{description}[style=unboxed, leftmargin=0cm]
\item[Tf-idf.]
This method calculates the tf-idf weight for each word from the input intents in the training data and the descriptions in the document set.

\item[Unsupervised NN.]
To encode intents and description, this method uses the pretrained model, \texttt{paraphrase-distilroberta-base-v1} in the Sentence-Transformers library \citep{reimers-gurevych-2019-sentence}.
This model is the distilled version of RoBERTa-base model \citep{sanh2019distilbert} with the mean pooling, which is trained with large-scale paraphrase data.

\item[Supervised NN.]
This method uses the fine-tuned \texttt{paraphrase-distilroberta-base-v1} with the training set of our data to vectorize intents and descriptions.
\end{description}
Please refer to Appendix B for the details of preprocessing and fine-tuning process.
In our experiments, we set $k$, the number of extracted signature-description pairs, to 5 for simplicity.

\begin{table}[t]
\centering
\begin{tabular}{l|cc}
\hline
                & Random & Library \\ \hline
Tf-idf          & 7.6                         & \textbf{18.89}              \\
Unsupervised NN & 9.6                         & 16.29                       \\
Supervised NN   & \textbf{37.1}               & 4.23                        \\ \hline
\end{tabular}
\caption{Recall@5 on the development set of the random and library split.}
\label{tab: retriver result}
\end{table}

We evaluate the performance of the retrievers with Recall@5 on the development set to chose the retrieving method.
Table \ref{tab: retriver result} shows the performance of the three methods.
While Supervised NN performs best in the random split, Tf-idf outperforms the others in the library split.
We select these models for each split, respectively.
For the library split, Supervised NN does not work well due to overfitting.
In contrast, the performance of Tf-idf and Unsupervised NN on the library split is much higher than on the random split.
This might be because of two reasons.
First, intents seem to well specify which third-party library should be used.
For example, given intents that include the word "array", a retriever can easily extract signatures of Numpy from the document set.
Second, our document set only contains the signature-description pairs of third-party libraries that appear in the data.
It makes retrieving easier because the number of signature-description pairs for each third-party library is small.
The construction of larger datasets that reflect more real situations will be future work.

Based on the document set and selected API retrievers, we evaluate our model in the following two settings:
\begin{description}[style=unboxed, leftmargin=0cm]
\item[Oracle.]
The extracted signature-description pairs $D^\prime$ consist only of the correct ones for each example. This setting measures how well our framework would work if the API retriever were perfect.
\item[Partially-Real-World.]
$D^\prime$ consists of pairs extracted by the API retriever. This setting investigates how much performance we can get with the current simple API retriever.
\end{description}

\subsection{Compared Models}
We use the CopyNet as a baseline that can copy symbols only from input intents.
This model can be regarded as a variant of our model in Section \ref{sec: Model} that does not have the API retriever or API reader, and includes only intent representations $\mathbf{h}^x_i$ in $H$.
We compare this baseline with our API reading models in Section \ref{sec: Model} that can refer to the extracted signature-descriptions $D^\prime$ from the selected API retrievers.
All models are trained with the cross-entropy loss.
Our preprocessing, neural network choice, and hyperparameter setting in the experiments are described in Appendix C.

\subsection{Results}

\begin{table}[]
\centering
\scalebox{0.9}{
\begin{tabular}{l|cc}
\hline
                          & \multicolumn{1}{l}{Random} & \multicolumn{1}{l}{Library} \\ \hline
Seq2seq\tablefootnote{\url{https://conala-corpus.github.io/}}                     & 10.58 & - \\
CopyNet                    & 21.7                      & 12.15                       \\ \hline
Reading (Oracle)           & 36.7                      & 15.98                       \\
Reading (P-Real-World)       & 21.88                      & 13.58                       \\ \hline
\end{tabular}
}
\caption{\label{tab: results}
BLEU scores the random and library splits of the CoNaLa dataset. Reading (P-Real-World) corresponds to our model in the partially real-world setting.
}
\end{table}

Table \ref{tab: results} shows the results on the random and library split of the CoNaLa dataset.
We can see that there is a large difference between the performance of the random and library split. This shows the difficulty of generating snippets using unknown libraries in the open-domain code generation.
In both splits, the proposed method in the oracle settings outperforms the baseline. This indicates that when the API retriever works properly, the framework of referencing API documents is effective for both known and unknown libraries.
In the partially real-world setting on the random split, the improvement from the baseline is marginal.
It seems to be because of noisy extraction from the document set.
On the other hand, the proposed method outperforms the baseline for the library split in both of the oracle and partially real-world settings.
This shows that the proposed model is effective when handling unknown libraries.
These results show that our implementation of the framework of the code generation with reading API documentations is effective when the API retriever works properly.

\section{Analysis}

\subsection{Recall Improvement}
To investigate whether the proposed method can properly generate unknown primitives that cannot be handled by existing methods, we calculate the recall of OOV primitives in the gold snippets, which is not included in the vocabulary V or the corresponding input intents.
Table \ref{tab: recall} shows the recall of OOV primitives on the development set in the library split.
As we expected, the baseline model of the ordinary supervised framework generated no OOV primitives.
In contrast, our models generated some of the OOV primitives from the API documentations.
Again, how well the OOV primitives are generated depends on the performance of the API retriever.
We can see that the more noise is included in $D^\prime$, the lower the recall is.

\begin{table}[]
\centering
\begin{tabular}{l|c}
\hline
                           & OOV Recall \\ \hline
CopyNet                    & 0.0        \\
Reading (Oracle)           & 21.6      \\
Reading (P-Real-World)       & 4.8        \\ \hline
\end{tabular}
\caption{\label{tab: recall}
Recall of the OOV primitives on the development set of the library split.
}
\end{table}

\begin{table}[!h]
\centering
\footnotesize
\scalebox{0.8}{
\begin{tabularx}{\linewidth}{lX}
\hline
Intent              & Get the integer location of a key 'bob' in a pandas data frame                   \\ \hline
\Checkmark                   & \code{df.index.\textbf{get\_loc}('bob')}                                                         \\
$\clubsuit$                   & \code{print(' '.join(map(str, bob)))}                                                   \\
$\heartsuit$                   & \code{"bob".\textbf{get\_loc}('bob')}                                                            \\
$\diamondsuit$                   & \code{\textcolor{gray}{\textbf{inspect}}\textbf{.get\_loc}(key, 'bob')}                                                     \\ \hline
... & \\
\multirow{2}{*}{2:} & \code{Index.\textbf{get\_loc}(key, method=None, tolerance=None)}                                 \\
                    & Get integer location, slice or boolean mask for requested label.                           \\
...                 &                                                                                  \\
\multirow{2}{*}{5:} & \code{\textcolor{gray}{\textbf{inspect}}.getouterframes(frame, context=1)}                                         \\
                    & Return an array of bytes representing an integer. The integer is ...                            \\ \hline \hline
Intent              & Rotate the xtick labels of matplotlib plot 'ax' by '45' degrees ...              \\ \hline
\Checkmark                   & \code{ax.\textbf{set\_xticklabels}( labels, rotation=45)}                                        \\
$\clubsuit$                   & \code{sorted(ax, 45)}                                                                   \\
$\heartsuit$                   & \code{ax.\textbf{set\_xticklabels}('ax')}                                                        \\
$\diamondsuit$                   & \code{\textcolor{gray}{\textbf{columns}}.\textcolor{gray}{\textbf{set\_yticklabels}}('ax', 45)}                                               \\ \hline
\multirow{2}{*}{1:} & \code{Axes.\textcolor{gray}{\textbf{set\_yticklabels}}(self, labels, ...)}                                         \\
                    & Set the y-tick labels with list of strings labels.                               \\
\multirow{2}{*}{2:} & \code{Axes.\textbf{set\_xticklabels}(self, labels, ...)}                                         \\
                    & Set the x-tick labels with list of string labels.                                \\
...                 &                                                                                  \\
\multirow{2}{*}{5:} & \code{DataFrame.\textcolor{gray}{\textbf{columns}}: Index}                                                         \\
                    & The column labels of the DataFrame.                                              \\ \hline \hline
Intent              & Sort array 'arr' in ascending order by values of the 3rd column                  \\ \hline
\Checkmark                   & \code{arr{[}arr{[}:, (2){]}.\textbf{argsort}(){]}}                                               \\
$\clubsuit$                   & \code{sorted(arr, key=lambda x: x{[}1{]})}                                              \\
$\heartsuit$                   & \code{arr = {[}line.\textbf{argsort}() for x in arr{]}}                                          \\
$\diamondsuit$                  & \code{arr.\textcolor{gray}{\textbf{order\_by}}(arr)}                                                               \\ \hline
...                 &                                                                                  \\
\multirow{2}{*}{5:} & \code{\textcolor{gray}{\textbf{order\_by}}(*fields)}                                                               \\
                    & By default, results returned by a QuerySet are ordered by the ordering tuple ... \\ \hline
\end{tabularx}}
\caption{\label{tab: examples}
Generated snippets and extracted signature-description pairs by the API retriever on the development set of the library split, where \Checkmark is the ground-truth and $\clubsuit$ is the baseline output. $\heartsuit$ and $\diamondsuit$ correspond to our model's outputs in the oracle and partially real-world setting, respectively.
Correct and erroneous copying from the signatures by our models is marked with \textbf{bold} and \textcolor{gray}{\textbf{gray-bold}} styles, respectively.
}
\end{table}

\subsection{Generated Examples}
Table \ref{tab: examples} shows the generated examples and the extracted signature-description pairs by the API retriever on the development set in the library split.

In the first example, our models copied the correct class method \code{get\_loc} from the extracted pairs.
In contrast, the baseline CopyNet could not generate this primitive because it is not included in the vocabulary constructed from the training data.
In the second example, although our oracle model copied the correct method \code{set\_xticklabels}, our model in the partially real-world setting copied the incorrect but similar one \code{set\_yticklabels}.
This example indicates that handling extraction noise and similar signatures is very important for our framework.
It needs more sophisticated API retriever and reader than the current simple ones.
The third example needs to recognize the library-specific slicing of Numpy.
This kind of library-specific notion is hard to cover in the ordinary supervised framework without training data.
It is also difficult for our current framework because signature-description pairs do not provide such information.
It seems that exploiting usage examples or tutorials in documentations is a promising direction.

\section{Related Works}

\subsection{Code Generation}
There are significant researches on developing neural models for code generation \citep{ling-etal-2016-latent, rabinovich-etal-2017-abstract, yin-neubig-2017-syntactic, dong-lapata-2018-coarse, iyer-etal-2018-mapping, yin-etal-2018-structvae, yin-neubig-2019-reranking, iyer-etal-2019-learning, xu-etal-2020-incorporating}.
For example, \citet{rabinovich-etal-2017-abstract} and  \citet{yin-neubig-2017-syntactic} proposed grammar-based neural models that generate an abstract syntax tree of the code.
These studies assume that training data cover all primitives. This assumption is not necessarily valid when we consider the open-domain code generation that handles unknown libraries like our setting.
In this ordinary supervised framework, we have to collect additional training data if we want to generate code for a newly released library.
In contrast, this paper investigates the framework using API documentations to support unknown libraries without additional training data, although we use simple encoder-decoder models with copy mechanism in our experiments.
We leave it as future work to incorporate these sophisticated models into our framework.

Similar to this paper, some studies used additional resources including API documentations for code generation.
\citet{ling-etal-2016-latent} proposed a code generation model for trading card games that can copy tokens from additional structured inputs such as card name and costs.
\citet{iyer-etal-2018-mapping} evaluated the generation of class member functions of JAVA by referring to class variables and methods in context.
While their curated dataset was split based on repositories to evaluate the setting that requires handling the out-of-domain environment variables and methods, our library-split setting requires handling the primitives of unknown libraries that do not appear in training data.
\citet{xu-etal-2020-incorporating} used signature-description pairs from the official Python document as a pretraining resource to solve limited data issues in the ordinary supervised framework.
Although pretraining on API documents might allow a model to generate unknown primitives that do not appear in the training data, it still requires retraining and finetuning for additional new libraries.
In contrast, this paper explores the model that can adapt unknown libraries without retraining by reading API documents.

\subsection{Zero-shot Semantic Parsing}
There are semantic parsing studies other than code generation that deal with out-of-domain settings.
In the Text2SQL task, which generates SQL statements given a natural language intent and table schema as inputs, researchers have developed models that can generalize to table schemas that did not appear during training \citep{yu-etal-2018-spider, zhong-etal-2020-grounded, Chang2020ZeroshotTL, wang-etal-2020-rat, suhr-etal-2020-exploring, wang2021meta}.
\citet{pasupat-liang-2015-compositional} proposed a semantic parsing model for question answering on unknown tables.
In code generation, unlike Text2SQL and table question answering, there are no additional structured inputs such as table schemes.
Therefore, our framework extracts relevant code signatures from a document set that contains thousands of signature-description pairs.

Some previous researches used natural language keywords assigned to formal representations.
\citet{herzig-berant-2018-decoupling} proposed a method separating constant prediction from structure prediction in generating formal representations.
They exploited a predefined lexicon mapping each constant into a natural language keyword to match between phrases in a natural language input and candidate constants for semantic parsing of knowledge-base question answering.
\citet{givoli-reichart-2019-zero} also proposed a method to deal with out-of-domain instruction execution tasks by combining a similar lexicon into a semantic parser.
Our framework differs in that we handle thousands of code signatures by retrieving relevant information like open-domain question answering, and that our model learns the alignment between each part of the code signature and the natural language descriptions since natural language keywords are not necessarily assigned to each part of code signatures in API documentations.

\section{Conclusion} \label{sec: Conclusion}
This paper tackled the important problem in the open-domain code generation, handling the unknown libraries.
We explored the framework that can refer to API documentations to adapt to unknown libraries without additional training like human programmers.
We implemented this framework with the retriever that can extract relevant code signatures based on the input intent, and the copy mechanism that can copy primitives in the extracted signatures.
Moreover, we extended the CoNaLa dataset and created the library split to evaluate code generation for unknown libraries and our framework.
Our experimental results showed that while the encoder-decoder model in the ordinary supervised framework could not generate unknown primitives, our framework could do.

In our future work, we will create a dataset that more approximates the real-world setting, where the document set contains all signature-description pairs for each library.
The other important direction is developing a model that can effectively exploit other useful information such as usage examples.

\bibliography{anthology,custom}
\bibliographystyle{acl_natbib}

\appendix
\section*{Appendix}
\label{appsec: hyperparameter setting}

\begin{table*}[]
\centering
\begin{tabular}{l|cccc}
\hline
                       & \multicolumn{4}{c}{Random}                        \\ \hline
                       & hidden size & dropout rate & document set & BLEU  \\ \hline
CopyNet                & 256         & 0.5          & -            & 25.57 \\
Reading (Oracle)       & 256         & 0.3          & -            & 40.82 \\
Reading (P-Real-World) & 384         & 0.3          & Oracle       & 25.52 \\ \hline \hline
                       & \multicolumn{4}{c}{Library}                       \\ \hline
                       & hidden size & dropout rate & document set & BLEU  \\ \hline
CopyNet                & 256         & 0.5          & -            & 16.11 \\
Reading (Oracle)       & 256         & 0.5          & -            & 20.67 \\
Reading (P-Real-World) & 128         & 0.5          & Oracle       & 17.11  \\ \hline
\end{tabular}
\caption{The best hyperparameters and BLEU scores in the random and library split}
\label{apptab: hyperparameters}
\end{table*}

\section{Disscussion in the Annotation}
\label{appsec: disscussion in the annotation}
In annotating pairs of code signatures and natural language descriptions,
we encountered some problematic cases as follows:
\begin{description}[style=unboxed, leftmargin=0cm]
\item[Not Found.]
The annotators could not find the description of the primitives. (34 examples)

\item[Renamed.]
The primitive in the snippet is renamed in the current document, such as \code{findALL} renamed as \code{find\_all} in BeautifulSoup  (2 examples)

\item[Referring to Other Libs.]
The primitive of a library in the snippet calls a function in other libaries, e.g. \code{scipy.matrix} calling \code{numpy.matrix} and there is no documentation for that. (5 examples)

\item[Not Python.]
The snippet is not Python based on the intent e.g. "\textit{Spawn a process to run python script `myscript.py` in C++}" (4 examples).
\end{description}

For the first case, Not Found, we did not annotate the signature-description pair for the primitive.
For the second Renamed case, we modified the primitive in the snippet to the new one.
For the third Referring to Other Libs case, the annotator extracts the signature-description pair of the called function in the other library.
For the last case, Not Python, we removed those examples from the dataset.

\section{API Retriever}
\label{appsec: api retriever}
For API retriever, we try to use the three methods, Tf-idf, Unsupervised NN, and Supervised NN.
We describe preprocessing process for Tf-idf and fine-tuning process for Supervised NN in the following sections.

\subsection{Preprocessing in Tf-idf}
To calculate the tf-idf weight for each feature, first, we use the Porter stemmer to stem tokens in the input intents in the training data and the descriptions in the document set.
Then we calculate tf-idf weights with the scikit-learn\footnote{\url{https://scikit-learn.org/stable/}} for unigram and bigram features.

\subsection{Finetuning in Supervised NN}
We finetune the distilled RoBERTa-base model on the training data and the annotated signature-description pairs with the sentence-transformers library\footnote{\url{https://www.sbert.net/}}.
For each instance, we construct the positive examples from the input intent with annotated descriptions, and the negative examples from the top 50 incorrectly extracted descriptions by the Tf-idf method.
Based on this data, we train the model and select the best model on the development set.
We set the batch size to 16 and the number of epochs to 5.
For the random split, we train the model with the contrastive loss to increase cosine similarities for the positive intent-description pairs and decrease for the negative pairs.
We set the warmup steps to 500 and use the default setting of the library for the other hyperparameters.
For the lexical split, where training is difficult, we tune hyperparameters more elaborately.
We explore the learning rate in $\{2e-5, 1e-5, 1e-6\}$.
We also try to train the model with the triplet loss.

\section{Preprocessing and Hyperparameters}
\label{appsec: hyperparameter setting}

\subsection{Preprocessing}
Code snippets are tokenized based on non-alphabet or non-numeric symbols.
We tokenize natural language intents and descriptions with the Spacy tokenizer from AllenNLP\footnote{\url{https://allennlp.org/}}.
Then we separate each token based on non-alphabet or non-numeric symbols.

\subsection{Hyperparameter Setting}
We use the pretrained 300-dimensional GloVe embeddings, \texttt{glove.840B.300d}, for natural language tokens and randomly initialized 256-dimensional embeddings for tokens of snippets and signatures.
We set the size of the output embeddings to 256.

We use single-layer BiLSTM encoders for each $\mathrm{Encoder_{SG}}$ and $\mathrm{Encoder_{NL}}$ and the single-layer LSTM decoder.
We tune the hidden size of BiLSTM encoders in $\{128, 256, 384\}$ and the dropout rate in $\{0.3, 0.5\}$
For the $\mathrm{Att_{doc}}$ and $\mathrm{Att_{dec}}$, we use the bilinear attention.

We use the beam search during decoding at the evalutions and set the beam size to $15$.
To optimize the models, we use the Adam with the learning rate $0.001$ and the weight decay $\lambda = 0.01$.
We set the batch size to $64$ and the max decoding steps to $60$.
We train the models for $100$ epochs and set the patience to $20$ epochs.

When training the API reading models in the partially real-world setting, we explore which to use the oracle document set or the partially real-world one.

Table \ref{apptab: hyperparameters} shows the best hyperparameters for each setting.

\end{document}